\documentclass[sigconf]{acmart}

\usepackage{booktabs} 
\usepackage{amsfonts}
\usepackage{amsmath}
\usepackage{multirow}

\setcopyright{none}

\acmConference[RecSys 2017 Poster Proceedings]{RecSys 2017 Posters}{August 27-31}{Como, Italy} 

\newcommand{\tepoch}{T$_{epoch}$}
\newcommand{\tconv}{T$_{conv}$}

\newcommand{\ndcg}{{\textsc{NDCG@30}}}
\newcommand{\pre}{{\textsc{P@5}}}
\newcommand{\rec}{{\textsc{R@30}}}

\newcommand{\lightfm}{\textsc{LightFM}}

\newcommand{\pop}{\textsc{POP}}
\newcommand{\warp}{\textsc{WARP}}
\newcommand{\awarp}{\textsc{A-WARP}}

\newcommand{\xing}{\textsf{XING}}
\newcommand{\yelp}{\textsf{Yelp}}

\newcommand{\movietwenty}{\textsf{ML-20m}}

\begin{document}

\title{WMRB: Learning to Rank in a Scalable Batch Training Approach}

\author{Kuan Liu}
\affiliation{
   \institution{Information Sciences Institute}
 University of Southern California
}
\email{kuanl@usc.edu}

\author{Prem Natarajan}
\affiliation{
   \institution{Information Sciences Institute}
 University of Southern California
}
\email{pnataraj@isi.edu}

\begin{abstract}
We propose a new learning to rank algorithm, named Weighted Margin-Rank Batch loss (WMRB), to extend the popular Weighted Approximate-Rank Pairwise loss (WARP). WMRB uses a new rank estimator and an efficient batch training algorithm. The approach allows more accurate item rank approximation and explicit utilization of parallel computation to accelerate training. In three item recommendation tasks, WMRB consistently outperforms WARP and other baselines. Moreover, WMRB shows clear time efficiency advantages as data scale increases. 

\end{abstract}

\vskip -0.4em



\maketitle

\vskip -0.4em

\section{Introduction}
Rank based learning algorithms have been widely applied to recommendation problems. One prominent example is Weighted Approximate-Rank Pairwise loss \cite{weston2010large}. It achieves a high precision on the \textit{top} of a predicted ranked list instead of an \textit{averaged} high precision over the entire list. However, it is not scalable to large item set in practice due to its intrinsic online learning fashion. 
In this work, we address the limitation by proposing a novel algorithm and empirically demonstrate its advantages in both accuracy and time efficiency.


\vskip -0.4em

\section{Background}
\textbf{Notation.} Let x denote a user, y an item, and \textbf{Y} the entire item set. $\textbf{y}_x$ denotes items interacted by user x. $\bar{\textbf{y}}_x\equiv\textbf{Y} \setminus \textbf{y}_x$ is the irrelevant item set. We omit subscript x when there is no ambiguity. $f_y(x)$ denotes the model score.~
The rank of item y is defined as 
\begin{equation}
\label{eq:rank}
r_y = rank_y(f,x,\textbf{y}) =\sum_{\bar{y}\in\bar{\textbf{y}}} \textbf{I}[f_y(x) \le f_{\bar{y}}(x)],
\end{equation}
where \textbf{I} is the indicator function. 
Finally, $|t|_+ \equiv max(t,0), t\in\mathbb{R}$.

\textbf{Weighted Approximate-Rank Pairwise loss}
(WARP) develops an \textit{online} approximation of item ranks. Its critical component is \textbf{an iterative sampling approximation procedure}:
For each user-item pair (x, y) , sample $y'\in\textbf{Y}$ uniformly with replacement until $1+f_{y'}(x)<f_y(x)$ is violated.
It estimates item ranks by

\begin{equation}
\label{eq:warp3}
r_y\approx rank^{warp}_y(f,x,\textbf{y}) = \lfloor \frac{|\textbf{Y}|-1}{N} \rfloor
\end{equation}
where N is the sampling times to find the violating example. It then incurs a rank-sensitive loss as in Order Weighted Average~\cite{yager1988ordered}, i.e.,

\begin{equation}
\label{eq:owa}
\Phi^{owa}(r_y) = \sum_{j=1}^{r_y} \alpha_j \quad \alpha_1\ge \alpha_2\ge..\ge 0,
\end{equation}
where the non-increasing $\alpha$ series control the sensitivity to ranks.

\vskip -0.4em

\section{Weighted Margin-Rank Batch loss}
\label{sec:owb}

\begin{figure}[t]
\vskip -1.0em
\centering
\caption{Standard deviations (relative values) of two types of rank estimators at different item ranks. Simulation is done with item set size N=100,000. `online' uses estimator (\ref{eq:warp3}) and `sampled-batch q' uses (\ref{eq:owb}) where $q=|\textbf{Z}|/|\textbf{Y}|$.}
\label{pic:std}
\vskip -0.3em
\includegraphics[width=0.89\columnwidth]{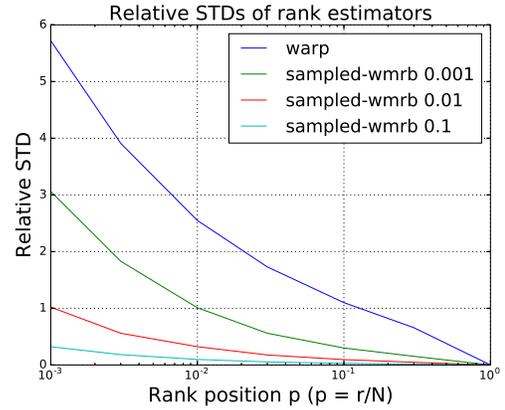}
\vskip -0.8em
\end{figure}

\textbf{Limitations of WARP}.
We first point out several limitations of WARP. 1) \textbf{Rank estimator} in (\ref{eq:warp3}) is not only biased\footnote{Suppose an item has a rank $r$ in a population N. Let $p = r / N$. Expectation of the estimator in (\ref{eq:warp3}) is approximately $p + \sum_{k=2}^N \frac{1}{k}p (1-p)^{k-1} > p$. It overestimates the rank seriously when r is small.} but also with \textbf{large variance}. As simulated in Fig.\ref{pic:std}, online estimation (blue) has a large relative variance, especially when items have high ranks (small $p$). 2) \textbf{Low updating frequency} results in prolonged training time in practice. This is because it can take a large number of sampling iterations to find a violating item, especially after the beginning stage of training. 3) \textbf{Intrinsic sequential manner} of WARP prevents full utilization of available parallel computation (e.g. GPUs), making it hard to train large or flexible models. 

\begin{table*}[!t]
\centering
\caption{Recommendation accuracy comparisons (in \%). Best results are in bold (e.g., \textbf{12.6}). WMRB outperforms pairwise based methods as well as batch based method CE.}
\vskip -1.3em
\begin{tabular}{|c|c||c|c|c||c|c|c||  c| c| c|} \hline
\multicolumn{2}{|c||}{\textbf{Datasets}}  & \multicolumn{3}{c||}{\xing} & \multicolumn{3}{|c||}{\yelp}  & \multicolumn{3}{|c|}{\movietwenty} \\ \hline
\multicolumn{2}{|c||}{\textbf{Metrics}}  & \pre  & \rec  & \ndcg  & \pre  & \rec    & \ndcg  & \pre  & \rec  & \ndcg  \\ \hline
- & \pop        &0.5&2.7&1.3  &0.3&0.9&0.5    &6.2&10.0&8.5 \\ \hline
\multirow{ 2}{*}{Pairwise}&\warp~['15]      &2.6&8.3&5.6  &1.3&4.4&2.5    &9.8&14.2&13.4 \\ \cline{2-11}
&\awarp~['15]     &2.6&11.6 &6.7  &1.3&4.3&2.5    &10.1&13.3&13.5 \\ \hline
\multirow{2}{*}{Batch}&CE ['16]      &2.5& 12.3 & 6.5 &1.4&4.5&2.6  &9.6&14.3&13.2 \\ \cline{2-11}
& WMRB       &\textbf{3.0}& \textbf{12.6}&\textbf{7.2}   &\textbf{1.5}&\textbf{5.1}& \textbf{2.9}  &\textbf{10.2}&\textbf{14.6}&\textbf{13.9} \\ \hline

\end{tabular}
\label{t:accuracy}
\vskip -1.2em
\end{table*}

\vskip -0.3em

\subsection{Proposed Approach}
We address the limitations by combining a rank-sensitive loss and batch training. We first define (sampled) \textbf{margin rank} as

\begin{equation}
\label{eq:owb}
rank^{wmrb}_y(f,x,\textbf{y}) =\frac{|\textbf{Y}|}{|\textbf{Z}|}\sum_{y'\in \textbf{Z}} |1-f_y(x) + f_{y'}(x)|_+ \textbf{I} (y'\in\bar{\textbf{y}}),
\end{equation}
where $\textbf{Z}$ is a subset of $\textbf{Y}$ randomly sampled (without replacement).

While \textit{margin rank} defined in (\ref{eq:owb}) is not the rank in (\ref{eq:rank}), it characterizes overall score violation of irrelevant items. The margin loss is often used to approximate the indicator function. Moreover, (\ref{eq:owb}) can be readily computed in a batch manner---Model scores between a user and a batch of items $f_{y'}(x)  \forall y' \in\textbf{Y}$ are first computed; The margin losses of violating items are then summed up.

We then design a \textbf{rank-sensitive loss function}. Note the \textit{margin rank} $r_y=rank^{wmrb}_y(f,x,\textbf{y})$ is a non-negative real number rather than an integer as in (\ref{eq:owa}). We define a differentiable loss function to incur ``rank weighted'' loss as follows:
\begin{equation}
\label{eq:owbloss}
L^{wmrb}(x,y) = \Phi^{wmrb}(r_y) = \log(r_y+1).
\end{equation}

By noting $\Phi''(r)=-\frac{1}{(1+r)^2}<0$, the loss is more sensitive with small r, thus mimicking the property as in (\ref{eq:owa}).

\paragraph{Compared to WARP}, WMRB replaces the sequential sampling procedure with batch computations. It results in a different rank approximation and loss function. Per user-item pair, WMRB involves more computation and is compensated with easy utilization of parallel computing. WMRB updates model parameters much more frequently than WARP -- which only updates the parameters of one user-item after many sampling.

WMRB has an unbiased estimator of margin rank. Its different sampling scheme results in smaller variances. Simulation in Fig.\ref{pic:std} shows sampled-wmrb has much smaller variance than \textit{warp} as long as $|\textbf{Z}|/|\textbf{Y}|$ is not too small.

\begin{table}
\centering
\caption{Dataset statistics. U: users; I: items; S: interactions.}
\vskip -0.9em
\label{t:data1}
\small{
\begin{tabular}{|c|c|c|c|c|}
\hline
\textbf{Data}   & $\left|U\right|$ & $\left|I\right|$ & $\left|S_{\text{train}}\right|$ & $\left|S_{\text{test}}\right|$ \\
\hline
\xing       & 1,500,000     & 327,002  & 2,338,766  & 484,237 \\ \hline
\yelp       & 1,029,433     & 144,073  &  1,917,218 & 213,460  \\ \hline
\movietwenty   & 138,493  & 27,278 & 7,899,901 & 2,039,972\\ \hline
\end{tabular}
}
\vskip -1.4em
\end{table}

\vskip -0.4em

\section{Results}
\label{sec:exp}

We validate WMRB on three datasets: XING\footnote{http://2016.recsyschallenge.com/}, Yelp\footnote{https://www.yelp.com/dataset\_challenge. Downloaded in Feb 17.}, and MovieLens-20m\footnote{https://grouplens.org/datasets/movielens/20m/}. The tasks are to recommend to users job posts, Yelp business, and movies, respectively. We assess the quality of recommendation by comparing models' recommendation to ground truth interactions split from the datasets. We report recommendation accuracies under metrics Precsion@5, Recall@30, and NDCG@30 as well as training time. The datasets statistics are listed in Table~\ref{t:data1}.

We compare WMRB to different methods.~\pop~recommends items purely based on popularity.~\warp~and~\awarp~are implemented in~\lightfm~\cite{kula_metadata_2015}.~\awarp~differs from vanilla WARP by incorporating available attributes. CE uses Cross-Entropy loss function and is a batch training based algorithm implemented by ourselves. CE and WMRB incorporate attributes as in~\awarp.

\paragraph{Accuracy}
Table~\ref{t:accuracy} reports accuracy comparisons of different models. We highlight two observations. First, WMRB consistently outperforms~\warp~and~\awarp. For instance, the relative improvements are 8.6\%, 18.6\%, and 9.8\% on Recall@30. Second, with both being batch based methods, WMRB wins over CE clearly, indicating the effectiveness of the rank-sensitive loss. 





\paragraph{Time efficiency} 
Table~\ref{t:time} reports dataset complexity and training time of different models. To measure complexity, we report the total number of model parameters and the average number of attributes per user-item pair. While we make an effort to speed up each method,\footnote{WMRB are implemented based on Tensorflow and run on a single GPU (GeForce GTX TITAN X).~\lightfm~runs asynchronous SGD on Cython with 5 cores.(Intel Xeon CPU 3.30GHz)} we are most interested in how running time changes with different data scales given fixed models and configurations. 

From Table \ref{t:time}, WMRB is slower than~\lightfm~on~\movietwenty. It catches up quickly on~\yelp, where data scale increases. On~\xing, which has the largest model size and complexity, WMRB is much faster. The trend clearly shows scalability advantages of batch training based WMRB.

\begin{table}[t]
\caption{Dataset complexities and training time comparisons. (~\tepoch: average epoch time;~\tconv: total training time.) With increasing data scales, WMRB shows time efficiency advantages over pairwise based implementation.}
\vskip -0.8em
\label{t:time}
\centering
\begin{tabular} { |c|| c | c | c | c | c| c|} \hline
\multirow{ 2}{*}{\textbf{Datasets}}  & \# of & \# of & \multicolumn{2}{|c|}{\lightfm}  & \multicolumn{2}{|c|}{WMRB}  \\ \cline{4-7}
 & \small {Param.} & \small{Attr.}& \tepoch & \tconv & \tepoch & \tconv \\ \hline
\movietwenty &4.6M & 11 & \textbf{7m}  & \textbf{1.2h}    & 22m  & 3.3 h \\ \hline
\yelp    &        9.3M & 19 & 10m & 5.0 h   & \textbf{9m}   & \textbf{3.9 h} \\ \hline
\xing    &        12.1M & 33 & 94m & 31.2h  & \textbf{24m} & \textbf{20.7h} \\ \hline
\end{tabular}
\vskip -1.0em
\end{table}


\vskip -0.4em

\section{Conclusion}
\label{sec:conclude}

In this work, we propose a simple yet effective approach to scale up learning to rank algorithm. It implements the idea of ordered weighted average loss in a batch training algorithm. Our preliminary empirical study shows its effectiveness.
\vskip -0.1em

\bibliographystyle{ACM-Reference-Format}
\bibliography{main} 
\end{document}